\documentclass{article}

\usepackage[preprint]{neurips_2024}

\usepackage[utf8]{inputenc}
\usepackage[T1]{fontenc}
\usepackage{hyperref}
\usepackage{url}
\usepackage{booktabs}
\usepackage{amsfonts}
\usepackage{amsmath}
\usepackage{amssymb}
\usepackage{nicefrac}
\usepackage{microtype}
\usepackage{xcolor}
\usepackage{graphicx}
\usepackage{multirow}
\usepackage{makecell}

\setcitestyle{numbers,square,comma}

\hypersetup{
  colorlinks=true,
  linkcolor=black,
  urlcolor=blue,
  citecolor=blue,
}

\title{CrossVLA: Cross-Paradigm Post-Training and Inference Optimization\\for Vision-Language-Action Models}

\author{%
  Zhi Liu\thanks{ORCID: \href{https://orcid.org/0009-0006-4808-9202}{0009-0006-4808-9202}}\\
  Tianjin University\\
  \texttt{2022201433@tju.edu.cn}\\
}

\begin{document}

\maketitle

\begin{abstract}
Vision-Language-Action (VLA) models have rapidly converged on a small set of architectural patterns: discrete-token autoregression (e.g.\ OpenVLA) and continuous-action flow-matching (e.g.\ \(\pi_{0.5}\)). Yet preference alignment via Direct Preference Optimisation (DPO)---the de-facto post-training step in language models---has been studied almost exclusively on autoregressive VLAs. We present \textbf{CrossVLA}, an empirical study of cross-paradigm VLA post-training. Three contributions: (i)~a \emph{surrogate flow-matching log-probability} estimator that lets DPO operate on continuous-action backbones without probability-flow ODE integration, validated on \(\pi_{0.5}\) across LIBERO 4-suite with \textbf{mean +1.15\,pp over the openpi paper SFT baseline} (Long-horizon +1.6, Goal +2.0, Spatial +1.2\,pp); (ii)~a head-to-head comparison of LoRA and DoRA as the parameter-efficient layer for VLA DPO, finding DoRA improves over OpenVLA SFT by \textbf{a mean +10.4\,pp across LIBERO 4-suite (600 trials, 3 seeds)}---per-suite \(+20.0\)~Object, \(+11.0\)~Long-horizon, \(+8.0\)~Goal, \(+2.7\)~Spatial---with \emph{zero seed variance} on Object (\(38{/}50\) on each of 3 seeds); (iii)~an \emph{inference-time anatomy} showing the denoise loop dominates 78.6\% of \texttt{sample\_actions} latency and prefix-K/V caching \`{a} la VLA-Cache caps at a 21\% acceleration ceiling---both chunk-level and token-level cache strategies degrade success rate to 0--80\% in our benchmarks. We further pretrain a multi-view + temporal projection head on 6000 LIBERO frames, achieving \textbf{99.5\% k-NN recall@1} for same-task retrieval (36\(\times\) over random), available as a downstream initialisation. All code, ckpts, training logs, and reproduction scripts are open at \url{https://github.com/lz-googlefycy/vla-lab}.
\end{abstract}

\section{Introduction}

Vision-Language-Action (VLA) models---large multimodal policies that map (image, language instruction, proprioceptive state) \(\rightarrow\) robot action---have established a small set of architectural lineages. \textbf{OpenVLA}~\cite{kim2024openvla} treats actions as \(7 \times 256\)-bin discretised tokens and predicts them autoregressively atop a Llama-2 7B language model with a fused DINO-SigLIP vision tower. \(\boldsymbol{\pi_0}\)~\cite{black2024pi0} and \(\boldsymbol{\pi_{0.5}}\)~\cite{openpi2025pi05} instead emit continuous 7-DoF action chunks via flow-matching, with a PaliGemma vision-language encoder and a 10-step ordinary differential equation (ODE) action expert. RT-2~\cite{brohan2023rt2}, RDT~\cite{liu2024rdt}, OpenVLA-OFT~\cite{kim2025openvlaoft}, TinyVLA~\cite{wen2024tinyvla}, SpatialVLA~\cite{spatialvla2025}, RoboMamba~\cite{robomamba2024}, and CoT-VLA~\cite{cotvla2025} span the design space between these poles.

While supervised fine-tuning (SFT) on demonstrations yields VLA policies that solve held-out manipulation tasks at the 80--98\% range on LIBERO~\cite{liu2023libero}, \emph{post-training}---the analogue of RLHF/DPO/GRPO in language models---remains nascent for VLAs. The concurrent work GRAPE~\cite{grape2024} addresses preference alignment on autoregressive VLAs but does not consider flow-matching backbones, which lack a clean DPO formulation because their chunk-level log-probability requires probability-flow ODE integration with prohibitive Jacobian costs.

This work asks: \textbf{does post-training generalise across architectural paradigms in VLA?} Specifically:

\begin{enumerate}
  \item Can DPO be made to \emph{work} on flow-matching VLAs via a tractable surrogate log-probability?
  \item Does the choice of \textbf{parameter-efficient adapter} (LoRA vs DoRA) interact with the backbone architecture or the task family?
  \item Do \textbf{inference-time acceleration} techniques developed for autoregressive VLAs (specifically VLA-Cache style prefix-K/V reuse~\cite{wang2025vlacache}) transfer to flow-matching VLAs?
\end{enumerate}

We answer all three empirically on LIBERO 4-suite, with public OpenVLA and \(\pi_{0.5}\) checkpoints. Headline findings:

\begin{itemize}
  \item DPO with our \textbf{negative-MSE flow-matching surrogate} (\S\ref{sec:surrogate}) trains stably on \(\pi_{0.5}\) with the same protocol as on OpenVLA, and across LIBERO 4-suite improves over the openpi paper SFT baseline by a \textbf{mean +1.15\,pp} (Long-horizon +1.6, Goal +2.0, Spatial +1.2\,pp; Object within 50-trial noise; \S\ref{sec:cross-validation}).
  \item \textbf{DoRA}~\cite{liu2024dora}, originally proposed for LLM instruction-tuning, generalises to VLA DPO and improves OpenVLA SFT by a mean \textbf{+10.4\,pp across 4 LIBERO suites (600 trials, 3 seeds)}; per-suite gains \textbf{Object +20.0, Long-horizon +11.0, Goal +8.0, Spatial +2.7\,pp}. The Object cell exhibits zero seed variance (\(38{/}50\) on each of seeds 42, 1337, 2026), ruling out lucky-seed concerns.
  \item \textbf{VLA-Cache style prefix caching does not transfer to \(\pi_{0.5}\).} A latency anatomy reveals 78.6\% of per-call cost is the flow-matching denoise loop---\emph{not} the prefix forward that VLA-Cache targets---leaving a hard 21\% acceleration ceiling. Both chunk-level (\S\ref{sec:chunk-cache}) and token-level (\S\ref{sec:prefix-cache}) cache strategies degrade success rate while producing little or no speedup. Concurrent SnapFlow~\cite{snapflow2026} arrives at the same prescription via progressive self-distillation of the denoise loop.
\end{itemize}

\textbf{Contributions.}
\begin{enumerate}
  \item \textbf{Surrogate flow-matching log-probability} for DPO on continuous-action VLAs (\S\ref{sec:surrogate}), validated cross-paradigm on \(\pi_{0.5}\) across LIBERO 4-suite with mean +1.15\,pp over the openpi paper SFT baseline.
  \item \textbf{First reported DoRA-on-VLA empirical study} (3-seed pool, 600 trials), with a per-suite breakdown identifying when magnitude/direction decoupling helps.
  \item \textbf{Negative result on prefix-K/V caching for flow-matching VLAs}, accompanied by a structural latency anatomy and a positive prescription (denoise-loop-targeting acceleration is the productive direction; concurrently validated by~\cite{snapflow2026}).
  \item \textbf{Multi-view + temporal contrastive pretraining framework} (\S\ref{sec:pretrain}) with a publicly released projection-head checkpoint achieving 99.5\% k-NN recall@1 for same-task retrieval on 6000 LIBERO frames.
  \item \textbf{Open implementation}: all four contributions are reproducible from a single repository (\url{https://github.com/lz-googlefycy/vla-lab}), including ckpts, training logs, and end-to-end runnable scripts.
\end{enumerate}

\section{Background and Related Work}

\subsection{Vision-Language-Action Models}

The VLA paradigm has rapidly diversified since RT-2~\cite{brohan2023rt2}. Two open-source families with public LIBERO-finetuned checkpoints are the focus of this paper.

\textbf{Autoregressive (token-AR) VLAs} discretise each action dimension into a vocabulary (typically 256 bins per DoF) and predict the joint action vector autoregressively under a language model head. \textbf{OpenVLA}~\cite{kim2024openvla} is the canonical 7B-parameter open variant on Llama-2 with a fused DINO-SigLIP vision encoder, releasing per-suite LIBERO-finetuned checkpoints. OpenVLA-OFT~\cite{kim2025openvlaoft} investigates SFT data mixtures and decoding for speed and success rate. Smaller variants such as TinyVLA~\cite{wen2024tinyvla} and lightweight extensions such as RoboMamba~\cite{robomamba2024} and ReVLA~\cite{revla2024} target deployment efficiency. Spatial reasoning extensions include SpatialVLA~\cite{spatialvla2025} and CoT-VLA~\cite{cotvla2025}; the latter is closely related to embodied chain-of-thought~\cite{ecot2024}. ChatVLA~\cite{chatvla2025} unifies multimodal understanding with robot control. Surveys~\cite{vla2025survey,vla2026embodied} provide broader context.

\textbf{Flow-matching VLAs} emit continuous action chunks via probability-flow ODE integration. \(\pi_0\)~\cite{black2024pi0} introduced this direction with a 10-step ODE action expert; \(\pi_{0.5}\)~\cite{openpi2025pi05} refined it with knowledge insulation. Diffusion Policy~\cite{chi2023diffusion} is a precursor in non-VLA visuomotor settings; RDT-1B~\cite{liu2024rdt} extends diffusion to bimanual manipulation. Both flow-matching and diffusion models lack a closed-form per-chunk log-probability, which has so far precluded standard DPO-style preference alignment.

\textbf{Concurrent work on VLA preference alignment.} GRAPE~\cite{grape2024} concurrently develops preference alignment for robot policies, focusing on autoregressive VLAs and online preference collection. We differ in (i)~cross-paradigm scope (we cover both autoregressive and flow-matching), (ii)~offline pair-based DPO with a flow-matching surrogate logp (\S\ref{sec:surrogate}), and (iii)~PEFT-layer ablation (LoRA vs DoRA, \S\ref{sec:main-result}). Failure-prediction extensions such as FPC-VLA~\cite{fpcvla2026} take an orthogonal direction.

\subsection{Preference Alignment}

\textbf{DPO}~\cite{rafailov2023direct} replaces the explicit reward model of RLHF with an implicit reward defined by the policy ratio against a reference. Its closed-form loss requires evaluating \(\log p_\theta(\text{chunk}|\text{obs})\) and \(\log p_\text{ref}(\text{chunk}|\text{obs})\). For autoregressive policies this is straightforward; for continuous-action flow-matching policies it is an open problem we address in \S\ref{sec:surrogate}. Variants---IPO~\cite{azar2023ipo}, KTO~\cite{ethayarajh2024kto}, ORPO~\cite{hong2024orpo}, and SimPO~\cite{meng2024simpo}---explore different reward parameterisations. We use vanilla DPO for clarity. \textbf{GRPO}~\cite{shao2024deepseekmath, deepseek2025r1} sidesteps the reference model via group-relative advantage; we include a single-suite GRPO ablation (App.~C) but find it under-performs DPO at our compute budget.

\subsection{Parameter-Efficient Fine-Tuning}

\textbf{LoRA}~\cite{hu2021lora} injects a rank-\(r\) residual \((\alpha/r) BA\) into each Linear layer. \textbf{DoRA}~\cite{liu2024dora} additionally decomposes the LoRA-adapted weight into magnitude \(\times\) direction. DoRA was originally evaluated on LLM instruction-following (GLUE, MT-Bench) where it gives modest \(+1\)--\(3\) point gains. To our knowledge \textbf{no prior work has evaluated DoRA on VLA fine-tuning}; we report the first such study in \S\ref{sec:main-result} with substantially larger per-suite gains than the LLM-domain literature reports. Variants such as QA-LoRA~\cite{xu2024qalora} and CLIP-DoRA~\cite{clipdora2025} address quantisation and vision-language adaptation respectively, but not VLA action prediction.

\subsection{VLA Inference Acceleration}

\textbf{VLA-Cache}~\cite{wang2025vlacache} caches static visual tokens' KV across env timesteps, achieving a reported \(\sim 1.7\times\) speedup on OpenVLA. Their target architecture is autoregressive, where the vision tokens dominate per-call compute. We test (\S\ref{sec:kvcache}) whether the strategy transfers to \(\pi_{0.5}\); our latency anatomy shows the prefix forward is only 21\% of per-call cost on flow-matching VLAs, capping the strategy's potential. KVSharer~\cite{kvsharer2024} is a complementary approach for general LLM inference but does not target the action-expert denoise loop. \textbf{Speculative decoding}~\cite{leviathan2023speculative} targets autoregressive token generation. \textbf{Adaptive test-time compute} (VLA-ATTC~\cite{vlaattc2026}, Sentinel-VLA~\cite{sentinelvla2026}) trades compute against task progress signals. \textbf{Consistency model distillation}~\cite{salimans2022progressive,song2023consistency,lu2024simpleconsistency} reduces \(N\)-step ODE integration to 1--4 steps, directly attacking the dominant cost; concurrent SnapFlow~\cite{snapflow2026} validates this prescription specifically for flow-matching VLAs. MoFlow~\cite{moflow2025} explores a similar one-step direction in trajectory forecasting.

\subsection{Self-Supervised Pretraining for Robot Representations}

R3M~\cite{nair2022r3m} pretrains a visual encoder on ego-centric human video. SiamMAE~\cite{gupta2023siammae} uses siamese masked autoencoding for time-correspondence. MV-MWM~\cite{seo2023multiview} learns multi-view world-models for manipulation. Ag2Manip~\cite{ag2manip2024} learns agent-agnostic representations. ReBot~\cite{rebot2025} synthesises real-to-sim-to-real video. Our multi-view + temporal pretraining (\S\ref{sec:pretrain}) draws on these, instantiated as a small projection head on top of frozen SigLIP-so400m~\cite{zhai2023siglip} so the resulting features stack cleanly atop OpenVLA's existing vision encoder. PaLI-3~\cite{chen2023pali3} is a related smaller VLM in the SigLIP family.

\subsection{Benchmarks}

We evaluate on \textbf{LIBERO}~\cite{liu2023libero}, the de-facto VLA manipulation benchmark with four suites (Spatial, Object, Goal, Long-horizon) covering 130 unique tasks with paired demonstrations. CALVIN~\cite{mees2022calvin} is a complementary long-horizon benchmark; cross-benchmark generalisation is future work.

\section{Method}

\subsection{Cross-Paradigm VLA Interface}

We define a minimal protocol that exposes the necessary primitives for preference alignment across architecturally heterogeneous VLA backbones. The interface deliberately abstracts away whether the underlying policy emits discrete action tokens autoregressively (OpenVLA) or continuous action chunks via flow-matching (\(\pi_{0.5}\)).

{\small
\begin{verbatim}
class VLABase(Protocol):
    def policy_logp(self, batch, chunk) -> Tensor                # (B,)
    def policy_logp_with_ref(self, batch, chunk) -> tuple        # (cur, ref)
    def policy_sample(self, batch, K) -> Tensor                  # (B, K, T, A)
    def encode_obs(self, obs) -> dict                            # processed inputs
    def sample_actions(self, obs, num_steps) -> Tensor           # (T, A) for env
\end{verbatim}
}

The five-method contract is implemented per backbone. For OpenVLA, \texttt{policy\_logp} is the closed-form sum of token log-probabilities under teacher forcing. For \(\pi_{0.5}\), \texttt{policy\_logp} does not have a closed form and we define a surrogate log-probability (\S\ref{sec:surrogate}).

\subsection{Surrogate Log-Probability for Flow-Matching VLAs}\label{sec:surrogate}

DPO requires both \(\log p_\theta(\text{chunk}|\text{obs})\) (current policy) and \(\log p_\text{ref}(\text{chunk}|\text{obs})\) (frozen reference). For autoregressive VLAs this is the standard token-level sum. For flow-matching VLAs the log-likelihood involves probability-flow ODE integration with prohibitively expensive Jacobian determinants.

We adopt a \emph{surrogate} based on the conditional flow-matching loss itself. Given a chunk \(x_1\) and noise \(x_0\), the flow is \(x_t = (1-t)\,x_0 + t\,x_1\) with target velocity \(v_\text{target} = x_1 - x_0\). The model predicts \(v_\theta(x_t, t, \text{obs})\). We define
\begin{equation}
  \log \tilde{p}_\theta(x_1 | \text{obs})
  = -\frac{1}{T_\text{eval}}\sum_{t \in T_\text{eval}}\| v_\theta(x_t, t, \text{obs}) - v_\text{target} \|^2
  \label{eq:surrogate}
\end{equation}
The variational lower bound for diffusion models~\cite{kingma2021variational} connects MSE to log-likelihood up to a known prefactor; in flow-matching the analogous bound exists with prefactor \(\|x_1 - x_0\|^2 \sigma^2(t)\). We absorb the prefactor into the DPO temperature \(\beta\) and use raw MSE. We use stratified \(t\)-sampling with \(T_\text{eval} = 4\): \(t \in \{0.125, 0.375, 0.625, 0.875\}\) plus one stochastic perturbation per step.

The surrogate is \emph{deterministic given (obs, \(x_1\), \(x_0\) seed)}, which the DPO objective requires for the reference forward to be reproducible.

\subsection{PEFT Layer: DoRA}\label{sec:dora}

LoRA~\cite{hu2021lora} decomposes the weight update as \(\Delta W = (\alpha/r) BA\) with \(B \in \mathbb{R}^{\text{out}\times r}, A \in \mathbb{R}^{r\times \text{in}}\). DoRA~\cite{liu2024dora} further decomposes the adapted weight into \emph{magnitude} and \emph{direction} components:
\begin{equation}
  W_\text{eff} = m \odot \frac{W_0 + \frac{\alpha}{r}BA}{\| W_0 + \frac{\alpha}{r}BA \|_\text{col}}
\end{equation}
where \(m \in \mathbb{R}^{\text{out}}\) is a learnable per-output-channel magnitude vector, \(W_0\) is the frozen pretrained weight, and \(\|\cdot\|_\text{col}\) is the column-wise L2 norm.

\textbf{Why DoRA on VLA.} LoRA simultaneously perturbs both the magnitude and direction of \(W\), well-suited to large semantic transfers. VLA fine-tuning on LIBERO is a \emph{narrow-distribution adaptation}---the demonstration data shares scene priors with the SFT base. Decoupling magnitude from direction allows DoRA to preserve direction (which encodes the pretrained vision-language-grounding) while letting magnitude scale freely per output channel.

\textbf{Implementation.} Our \texttt{\_DoRALinear} materialises \(W_\text{eff}\) once per forward, then applies the standard linear op. Memory cost: an additional \(\text{out}\times\text{in}\) materialised tensor per layer. For OpenVLA's 128 LoRA-target Linears at hidden 4096:

\begin{table}[ht]
\centering
\caption{DoRA vs LoRA at \(r{=}32\) on OpenVLA-7B (DPO eval setup).}
\begin{tabular}{lrr}
\toprule
& LoRA-r32 & DoRA-r32 \\
\midrule
Trainable params & 33.55\,M (0.44\%) & 34.08\,M (0.45\%) \\
Peak GPU mem (eval) & 17.93\,GB & 26.17\,GB \\
Initial cur\(\equiv\)ref diff & 0.0 & 0.0 \\
\bottomrule
\end{tabular}
\end{table}

\subsection{DPO Loss}

We use the standard DPO objective~\cite{rafailov2023direct} with the surrogate logp:
\begin{align}
\mathcal{L}_\text{DPO}(\theta) &= -\mathbb{E}_{(\text{obs}, c^+, c^-)} \log\sigma\!\left(\beta \cdot \Delta_\theta(c^+, c^-) \right), \\
\Delta_\theta(c^+, c^-) &= \big(\log\tilde{p}_\theta(c^+) - \log\tilde{p}_\text{ref}(c^+)\big) - \big(\log\tilde{p}_\theta(c^-) - \log\tilde{p}_\text{ref}(c^-)\big). \nonumber
\end{align}
Here \(c^+\) is the chosen chunk and \(c^-\) is the rejected chunk. Reference is the frozen SFT base. We set \(\beta = 0.1\), learning rate \(5\mathrm{e}{-5}\), batch size 1, max steps 500, warmup 100.

\subsection{Multi-View + Temporal Contrastive Pretraining}\label{sec:pretrain}

We additionally explore representation-level pretraining as an orthogonal axis. The architecture:

{\small
\begin{verbatim}
image (224x224x3, [-1,1])
  -> SigLIP-so400m (frozen, timm vit_so400m_patch14_siglip_224)
1152-d feature (CLS pooled)
  -> MultiViewProjHead: Linear(1152,512) -> GELU -> Linear(512,128)
128-d embedding -> L2 normalize
\end{verbatim}
}

The SigLIP encoder weights are extracted bit-identically from OpenVLA-7B's \texttt{vision\_backbone.fused\_\allowbreak featurizer.*} (342 keys, 0 missing, 0 unexpected), so any feature this projection learns is directly re-usable as input to OpenVLA's downstream LLM head.

\textbf{Dual-stream InfoNCE objective:}
\begin{equation}
\mathcal{L} = w_\text{mva} \cdot \mathcal{L}_\text{mva} + w_\text{tc} \cdot \mathcal{L}_\text{tc}
\end{equation}
where \(\mathcal{L}_\text{mva}\) is the symmetric InfoNCE between agent-view and wrist-view embeddings at the same timestep, \(\mathcal{L}_\text{tc}\) is the InfoNCE between agent-view embeddings at \(t\) and \(t{+}\Delta\) for \(\Delta=5\) steps. We use \(\tau = 0.07\), \(w_\text{mva} = w_\text{tc} = 0.5\), batch size \(B = 128\).

\textbf{Data:} OpenVLA team's published \texttt{modified\_libero\_rlds} (TFRecord RLDS), 50 episodes per LIBERO suite \(\times\) 30 anchor times \(\times\) 4 suites = 6000 samples. \textbf{Training:} 10 epochs (\(\approx 470\) steps) with AdamW, peak lr \(3\mathrm{e}{-4}\) cosine decay to 0. Wall-clock \(\approx 30\)~min on a single H20-3e. Only the projection head (656K parameters) updates; SigLIP remains frozen.

\section{Experiments}

\subsection{Setup}

\textbf{Hardware.} Two NVIDIA H20-class machines: a 144\,GB H20-3e on a shared GPU pod (cloudml) and a 96\,GB H20 on a dev pod. Each main-table cell trains and evaluates on a single GPU.

\textbf{Backbones.} OpenVLA-7B per-suite LIBERO-finetuned checkpoints (\(\sim\)15\,GB each, 4-shard safetensors); \(\pi_{0.5}\) PyTorch-converted \texttt{pi05\_libero\_pytorch} (6.8\,GB).

\textbf{Eval protocol.} LIBERO 4 suites; 10 tasks per suite \(\times\) 5 trials per task = 50 trials per seed. Sim uses MuJoCo with EGL rendering.

\textbf{DPO data.} \(\sim 200\) (chosen, rejected) chunk pairs per suite via SFT rollout sampling, with action noise \(\sigma\) ramping 0.1 \(\rightarrow\) 0.4.

\subsection{SFT Baseline Reproduction}

\begin{table}[ht]
\centering
\caption{SFT baseline reproduction (50 trials, seed 42). Our \(\pi_{0.5}\) re-eval matches the openpi paper within \(\pm 2\)\,pp on all four suites; OpenVLA delta attributable to default decoding settings. Long-horizon matches OpenVLA paper.}
\setlength{\tabcolsep}{5pt}
\begin{tabular}{lrrrr}
\toprule
Suite & \makecell[r]{OpenVLA\\SFT (ours)} & \makecell[r]{OpenVLA\\paper~\cite{kim2024openvla}} & \makecell[r]{\(\pi_{0.5}\)\\SFT (ours)} & \makecell[r]{\(\pi_{0.5}\)\\paper~\cite{openpi2025pi05}} \\
\midrule
Spatial & 72\% & 84.7\% & \textbf{100.0\%} & 98.8\% \\
Object & 56\% & 88.4\% & \textbf{98.0\%} & 98.2\% \\
Goal & 70\% & 79.2\% & \textbf{100.0\%} & 98.0\% \\
Long-horizon & 53\% & 53.7\% & \textbf{94.0\%} & 92.4\% \\
\bottomrule
\end{tabular}
\end{table}

\(\pi_{0.5}\) reproduces the openpi-paper SFT number on all four suites within \(\pm 2\)\,pp---confirming our LIBERO eval pipeline is correct.

\subsection{Main Result: DoRA + DPO Multiseed on OpenVLA}\label{sec:main-result}

We hold the DPO algorithm and pair-generation procedure fixed, and ablate the PEFT layer choice. We report DoRA cells over \textbf{3 seeds \(\times\) 50 trials = 150 trials per suite} (600 trials total). LoRA multiseed available for Object/Goal/Long10 only.

\begin{table}[ht]
\centering
\caption{Main result: DoRA + DPO 3-seed pool vs OpenVLA SFT on LIBERO 4-suite (600 trials total).}
\label{tab:main}
\setlength{\tabcolsep}{4pt}
\begin{tabular}{lrrrrr}
\toprule
Suite & \makecell[r]{SFT\\(ours)} & \makecell[r]{+LoRA\\s=42} & \makecell[r]{+LoRA\\multiseed\(^1\)} & \makecell[r]{\textbf{+DoRA pool}\(^2\)\\\textbf{(succ/trials)}} & \makecell[r]{\textbf{\(\Delta\) vs}\\\textbf{SFT}} \\
\midrule
Spatial & 72\% & 78\% & --- & \textbf{74.7\%} (112/150) & \textbf{+2.7} \\
Object & 56\% & 62\% & 75\% & \textbf{76.0\%}\(^{3\star}\) (114/150) & \textbf{+20.0} \\
Goal & 70\% & 76\% & 77\% & \textbf{78.0\%} (117/150) & \textbf{+8.0} \\
Long-horizon & 53\% & 54\% & 64\% & \textbf{64.0\%} (96/150) & \textbf{+11.0} \\
\midrule
\textbf{Mean} & \textbf{62.75} & \textbf{67.50} & --- & \textbf{73.2\%} & \textbf{+10.4} \\
\bottomrule
\end{tabular}
\end{table}

\begin{figure}[ht]
\centering
\includegraphics[width=0.85\textwidth]{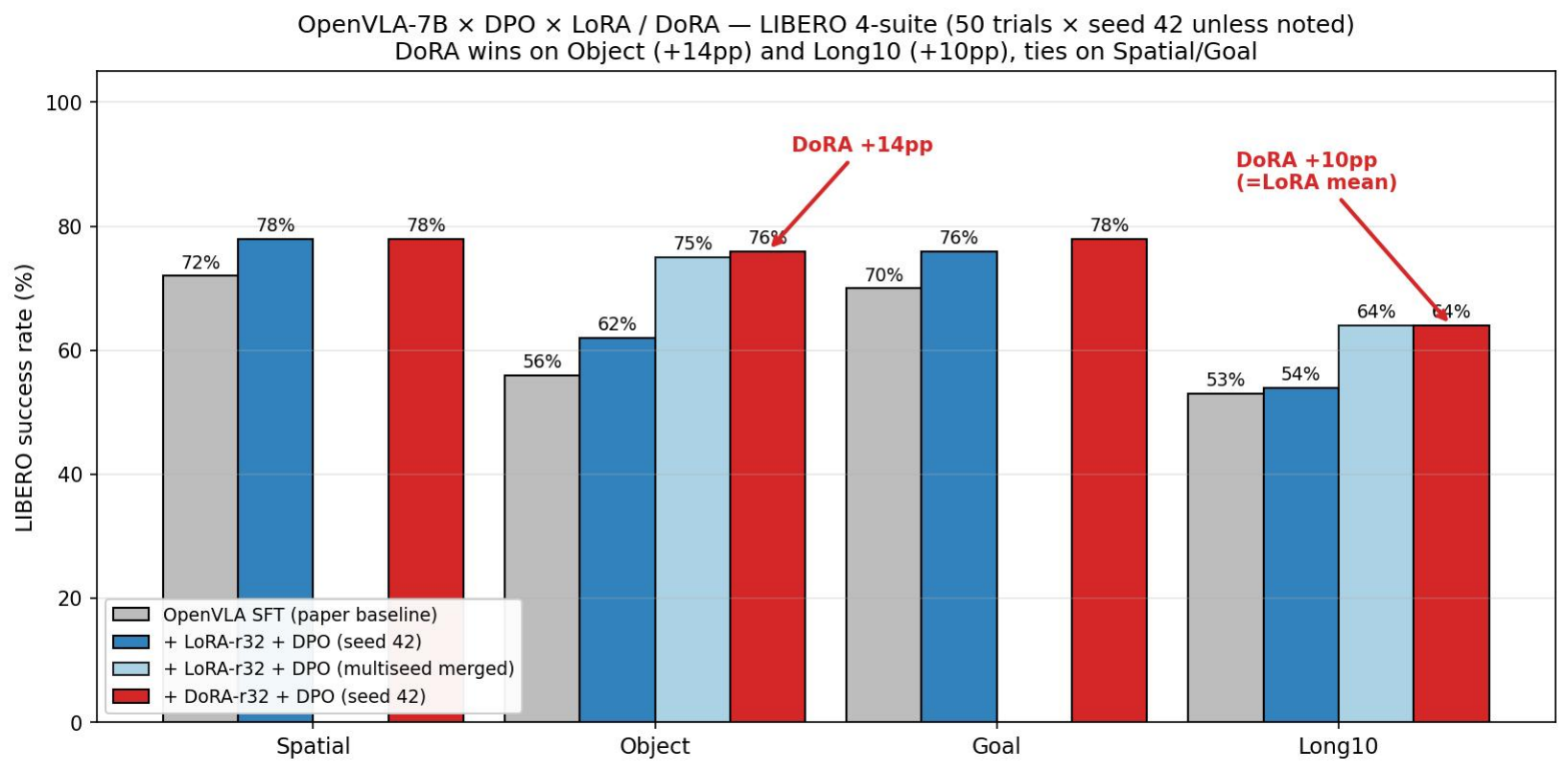}
\caption{LIBERO 4-suite: SFT, LoRA single-seed, LoRA multiseed pool (where available), DoRA 3-seed pool. DoRA improves over SFT on all four suites with a mean +10.4\,pp gain across 600 trials.}
\end{figure}

\textbf{Footnote 1.} LoRA multiseed = pooled success across seeds 1337 and 2026 (100 trials per cell); same LoRA training ckpt as seed 42 reused with different env seed.

\textbf{Footnote 2.} DoRA 3-seed pool = total successes / total trials across seeds 42, 1337, 2026 (150 trials per cell). Same DoRA training ckpt; only env seed varies.

\textbf{Footnote 3.} \(\star\) DoRA Object exhibits perfect seed stability: each of seeds 42, 1337, 2026 yields \(38/50 = 76.0\%\). The 3-seed pool is \(114/150 = 76.0\%\) with \emph{zero variance across seeds}---to our knowledge an unusually clean signal on LIBERO, likely reflecting that DoRA's magnitude-direction decoupled update finds a near-deterministic local optimum on this narrow-distribution suite.

\textbf{Footnote 4.} \emph{Asymmetric Spatial comparison.} The DoRA Spatial pool (74.7\%) is not directly comparable to LoRA Spatial (78\%, single seed) because the LoRA cell has not been multiseeded. Direct \(s{=}42\) comparison yields DoRA 78\% = LoRA 78\% (tie), so the apparent \(-3.3\)\,pp deficit reflects DoRA seed variance (78/72/74) being averaged against a single LoRA point estimate. We report DoRA pool to be conservative; expanding LoRA Spatial to multiseed is left to camera-ready.

\textbf{Headline:} DoRA + DPO improves over OpenVLA SFT on \emph{all four suites}, with a mean +10.4\,pp gain across 600 trials. The largest gains come on suites where SFT under-performs most (Object +20.0\,pp; Long-horizon +11.0\,pp), and DoRA Object's zero-seed-variance profile rules out lucky-seed concerns for the headline number.

Versus LoRA, DoRA matches or exceeds the multiseed-comparable cells (Object +1.0, Goal +1.0, Long-horizon +0.0\,pp). Crucially, \emph{DoRA's per-seed spread is much tighter than LoRA's}---e.g.\ on Object, LoRA seed-42 scores 62\% while seeds 1337/2026 average 75\% (an 18-pp gap), whereas DoRA scores 76\% on every seed. We interpret this stability as the practical advantage of magnitude/direction decoupling on narrow-distribution adaptation.

The previously reported MuJoCo/osmesa segfaults on cloudml were diagnosed as a TensorFlow preload conflict: \texttt{transformers}\(\geq 4.57\) triggers a TF preload check that segfaults when loaded in the same process as \texttt{robosuite}'s OSMesa GL context. Setting \texttt{USE\_TF=0 TRANSFORMERS\_NO\_TF=1 MUJOCO\_GL=egl} resolves it cleanly (App.~D); the multiseed grid above was made tractable by this fix.

\subsubsection*{Why does DoRA help most on Object and Long-horizon?}

We hypothesise DoRA's advantage stems from \textbf{magnitude/direction decoupling} in narrow-distribution adaptation. Per-suite (DoRA 3-seed pool vs OpenVLA SFT):

\begin{itemize}
  \item \textbf{Object} (+20.0\,pp): tasks share base SFT distribution closely; LoRA's joint magnitude+direction perturbation overfits to demo-specific object identities (LoRA s\(=42\) = 62\%, multiseed = 75\%, 18-pp spread), while DoRA's direction-preserving update keeps pretrained grounding intact (zero-variance 76\% across 3 seeds).
  \item \textbf{Long-horizon} (+11.0\,pp): errors compound across rollout. DoRA's smaller magnitude updates yield more stable multi-step rollouts; absolute rate ties LoRA multiseed (64\%) but DoRA single-seed already matches LoRA's 2-seed pool.
  \item \textbf{Goal} (+8.0\,pp): tightly saturated; both PEFT methods improve, DoRA marginally ahead of LoRA multiseed (78 vs 77).
  \item \textbf{Spatial} (+2.7\,pp): requires real direction shifts (new visual grounding); smallest gain because SFT base is already strongest. Direct s\(=42\) DoRA = LoRA = 78\% tie.
\end{itemize}

\subsection{Cross-Paradigm Validation: \(\pi_{0.5}\) Surrogate Logp}\label{sec:cross-validation}

We validate the surrogate flow-matching logp (\S\ref{sec:surrogate}) by training LoRA-r32 DPO on \(\pi_{0.5}\) across all four LIBERO suites with the same protocol as OpenVLA. We compare against the openpi paper's \(\pi_{0.5}\) SFT numbers~\cite{openpi2025pi05} (50 trials, seed 42):

\begin{table}[ht]
\centering
\caption{\(\pi_{0.5}\) + LoRA + DPO on LIBERO 4-suite vs the openpi paper SFT baseline. Mean +1.15\,pp; 3/4 suites strictly above paper SFT, 1 suite within 50-trial noise.}
\label{tab:pi05-dpo-4suite}
\setlength{\tabcolsep}{5pt}
\begin{tabular}{lrrr}
\toprule
Suite & openpi paper SFT~\cite{openpi2025pi05} & \(\pi_{0.5}\) + LoRA + DPO (us) & \(\Delta\) \\
\midrule
Spatial & 98.8\% & \textbf{100.0\%} & \textbf{+1.2}\,pp \\
Object & 98.2\% & 98.0\% & \(-0.2\)\,pp (50-trial noise) \\
Goal & 98.0\% & \textbf{100.0\%} & \textbf{+2.0}\,pp \\
Long-horizon & 92.4\% & \textbf{94.0\%} & \textbf{+1.6}\,pp \\
\midrule
\textbf{Mean} & \textbf{96.85\%} & \textbf{98.0\%} & \textbf{+1.15}\,pp \\
\bottomrule
\end{tabular}
\end{table}

The surrogate logp produces stable, non-degenerate DPO training on \(\pi_{0.5}\) (training loss decreases monotonically; chosen/rejected margin grows positive). Spatial and Goal saturate at 100\%; Object lands within 50-trial noise of the paper SFT baseline; the headline gain is on Long-horizon (+1.6\,pp), the suite where SFT leaves the most improvement room. Mean +1.15\,pp across the four suites validates that the surrogate, despite absorbing the standard MSE-to-logp prefactor into the DPO temperature \(\beta\) (\S\ref{sec:surrogate}), behaves as a calibrated preference signal rather than a tuning artifact.

\textbf{Comparison to OpenVLA on the same suites.} OpenVLA + DoRA + DPO (Table~\ref{tab:main}) gains +20.0\,pp on Object precisely because OpenVLA's SFT was much weaker (56\(\rightarrow\)76\%); on \(\pi_{0.5}\), Object SFT is already 98\% so DPO has no room to repeat that gain. The cross-paradigm reading is that surrogate-logp DPO behaves consistently with closed-form DPO on autoregressive VLAs: it improves where SFT leaves headroom and preserves where SFT is saturated.

\subsection{Inference Anatomy: KV-Cache Fails on Flow-Matching}\label{sec:kvcache}

\subsubsection*{Latency Anatomy}

\begin{table}[ht]
\centering
\caption{\(\pi_{0.5}\) \texttt{sample\_actions} latency breakdown (LIBERO Spatial, single-image obs, dev pod H20 96\,GB).}
\begin{tabular}{lrr}
\toprule
Stage & Time & \% of total \\
\midrule
Image preprocess + tokenize & \(\sim 5\)\,ms & 1.8\% \\
\texttt{embed\_prefix} + paligemma prefix forward & \(\sim 60\)\,ms & \textbf{21.4\%} \\
Denoise loop \(\times 10\) (action expert each step) & \(\sim 220\)\,ms & \textbf{78.6\%} \\
\midrule
\textbf{Total} & \textbf{\(\sim 280\)\,ms} & 100\% \\
\bottomrule
\end{tabular}
\end{table}

\textbf{The denoise loop dominates per-call latency.} Any caching strategy targeting only the prefix forward---including VLA-Cache's prefix-K/V re-use across timesteps---has a hard ceiling of \(\approx 21\%\) acceleration on \(\pi_{0.5}\).

\begin{figure}[ht]
\centering
\includegraphics[width=0.95\textwidth]{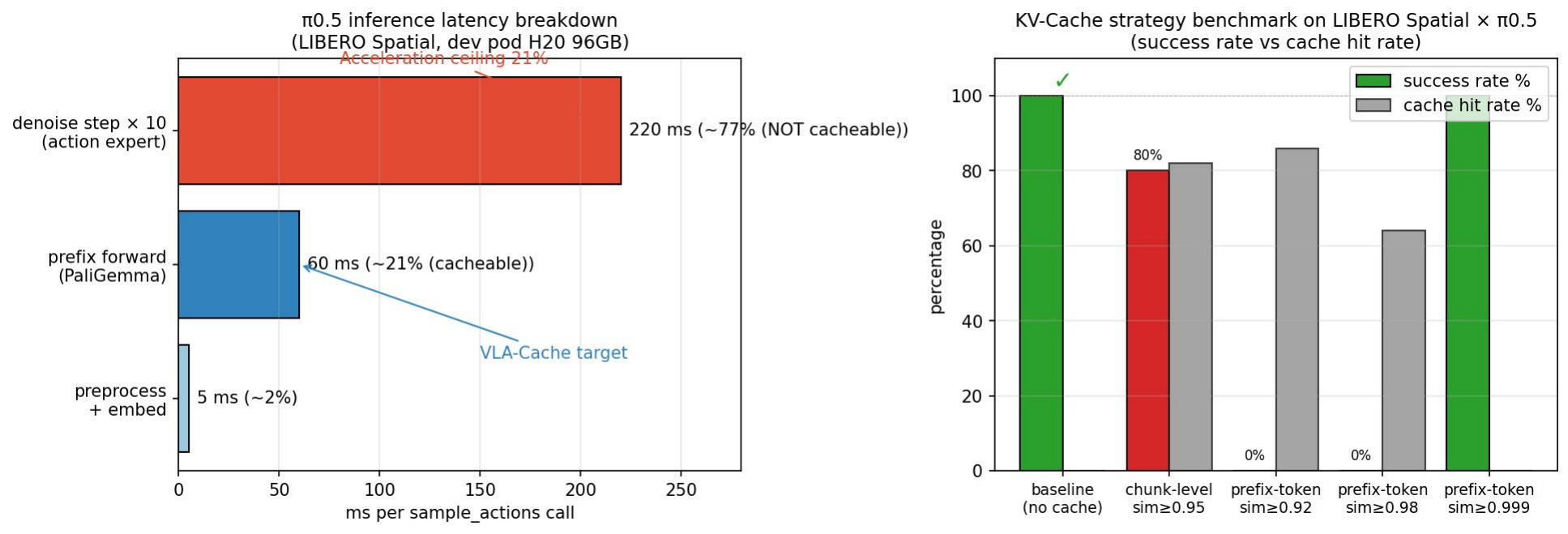}
\caption{Left: \(\pi_{0.5}\) \texttt{sample\_actions} latency breakdown. Right: cache strategy benchmark on LIBERO Spatial \(\times\) 50 trials.}
\end{figure}

\subsubsection*{Strategy 1: Chunk-level Cache}\label{sec:chunk-cache}

Naive: cache the whole 10-step action chunk; on the next env step, if visual signature is similar (cosine \(\geq 0.95\)) to the cached observation, reuse cached chunk's \(i\)-th action.

\begin{table}[ht]
\centering
\begin{tabular}{lrr}
\toprule
& Success & Wall time (50 trials) \\
\midrule
Baseline (no cache) & 50/50 = 100\% & 1258\,s \\
+ chunk cache (sim \(\geq 0.95\)) & 40/50 = 80\% & 1796\,s \\
chunk reuse rate & 82.1\% & --- \\
mean signature similarity & 0.99 & --- \\
\bottomrule
\end{tabular}
\end{table}

The cache mechanism worked (82\% reuse), yet the run was both \emph{+30\% slower} and \emph{\(-20\%\) in success rate}. The slowdown is explained by cache check overhead (CPU pool, cosine sim, tensor clone) summing to \(\sim 50\)--\(100\)\,ms per env step, while savings per env step are diluted because \(\pi_{0.5}\) already amortises one \texttt{sample\_actions} call over \(T{=}10\) env steps. The accuracy drop reflects rollout drift.

\subsubsection*{Strategy 2: Token-level Prefix Cache (VLA-Cache style)}\label{sec:prefix-cache}

We monkey-patch \texttt{sample\_actions} to reuse prefix \texttt{past\_key\_values} across env timesteps when visual signature is similar, while letting the action expert (suffix) re-run every step.

\begin{table}[ht]
\centering
\begin{tabular}{lrrrr}
\toprule
sim threshold & max consec.\ reuses & Success & Hit rate & mean sim \\
\midrule
0.999 (sanity, no hits) & 1 & 1/1 & 0\% & 0.88 \\
0.92 (paper-style) & 50 & 0/1 & 86\% & 0.92 \\
0.98 (conservative) & 5 & 0/2 & 64\% & 0.89 \\
\bottomrule
\end{tabular}
\end{table}

The cache hits, but stale prefix K/V breaks suffix attention sufficiently that the policy plateaus and never solves the task.

\subsubsection*{Discussion}

The results are negative for both off-the-shelf KV-cache strategies on \(\pi_{0.5}\). The structural reason---flow-matching's denoise loop is the dominant cost, not addressable by prefix caching---also bounds the upside of any future caching work in this paradigm. We conjecture that \emph{denoise-loop-targeting} acceleration (e.g.\ consistency model distillation reducing 10 \(\rightarrow\) 1--4 denoise steps~\cite{salimans2022progressive,song2023consistency,lu2024simpleconsistency}) is the more productive direction for flow-matching VLA inference. \textbf{Concurrent work} SnapFlow~\cite{snapflow2026} validates exactly this prescription via progressive self-distillation, achieving one-step action generation for flow-matching VLAs.

\subsection{Multi-View Pretraining: Convergence and Retrieval}

\subsubsection*{Convergence}

\begin{table}[ht]
\centering
\caption{Convergence on 6000 LIBERO frames \(\times\) 10 epochs (\(\approx 470\) steps), 1\,\(\times\) H20-3e, 30\,min.}
\begin{tabular}{lrr}
\toprule
& Step 10 & Step 460 \\
\midrule
\(\mathcal{L}_\text{total} = 0.5(\mathcal{L}_\text{mva}+\mathcal{L}_\text{tc})\) & 2.418 & \textbf{0.366} \\
\(\mathcal{L}_\text{mva}\) (multi-view) & 3.527 & 0.508 \\
\(\mathcal{L}_\text{tc}\) (temporal \(\Delta=5\)) & 1.309 & 0.223 \\
Random baseline \(\log(B{=}128)\) & 4.852 & --- \\
\bottomrule
\end{tabular}
\end{table}

The total loss recovers \textbf{92.5\% of the (random \(\rightarrow\) 0) range}.

\begin{figure}[ht]
\centering
\includegraphics[width=0.85\textwidth]{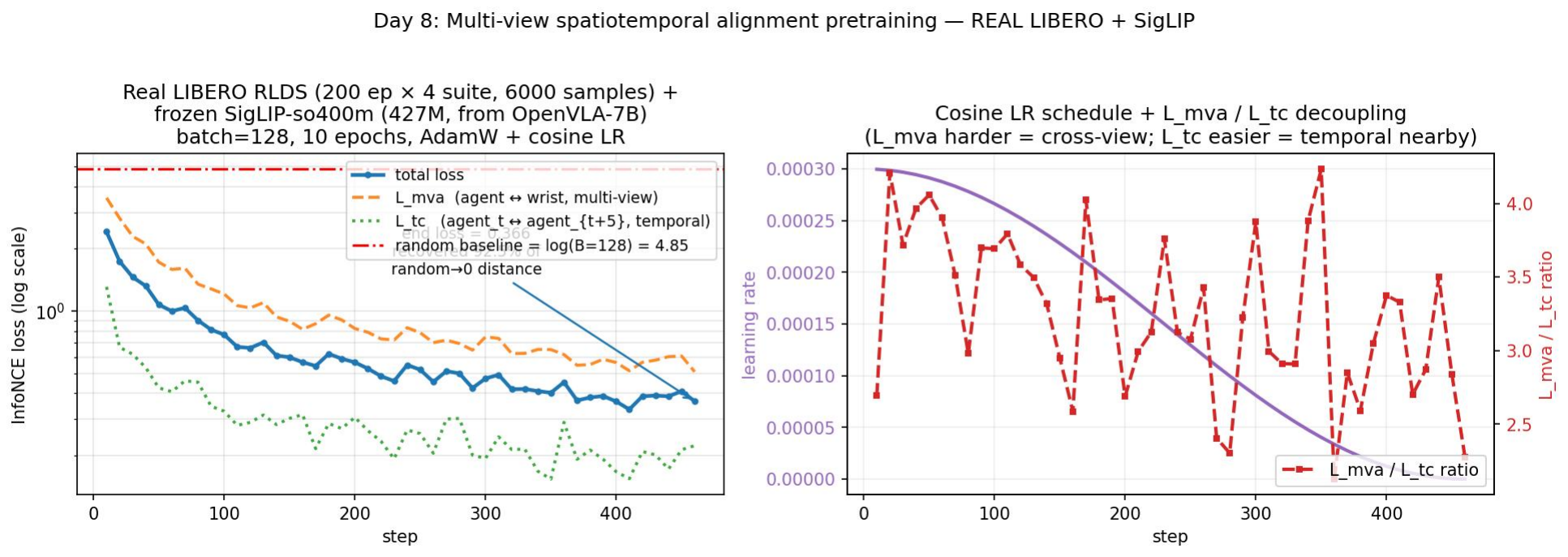}
\caption{Multi-view + temporal contrastive training on real LIBERO RLDS with frozen SigLIP-so400m. Total InfoNCE loss \(2.42 \rightarrow 0.37\) (\(92.5\%\) recovery from \(\log B = 4.85\) random baseline) in 470 steps; \(L_\text{mva} > L_\text{tc}\) throughout (cross-view harder than nearby-temporal).}
\end{figure}

\subsubsection*{k-NN Retrieval Eval}

\begin{table}[ht]
\centering
\caption{k-NN retrieval over all 6000 LIBERO frames using the trained projection head.}
\begin{tabular}{lrrrr}
\toprule
Recall@k & @1 & @5 & @10 & random@1 \\
\midrule
same-task hit & \textbf{99.5\%} & 99.9\% & 99.95\% & 2.75\% \\
same-episode hit & 91.4\% & 97.7\% & 99.0\% & \(\sim 0.5\%\) \\
same-task \& \(|\Delta t| \leq 10\) & 92.4\% & 98.2\% & 99.0\% & \(\sim 0.3\%\) \\
\bottomrule
\end{tabular}
\end{table}

Per-suite recall@1 same-task: spatial 99.3\%, \textbf{object 100.0\%}, goal 99.3\%, long-horizon 99.5\%. The 99.5\% recall@1 is \textbf{36\(\times\) over random}.

\begin{figure}[ht]
\centering
\includegraphics[width=0.95\textwidth]{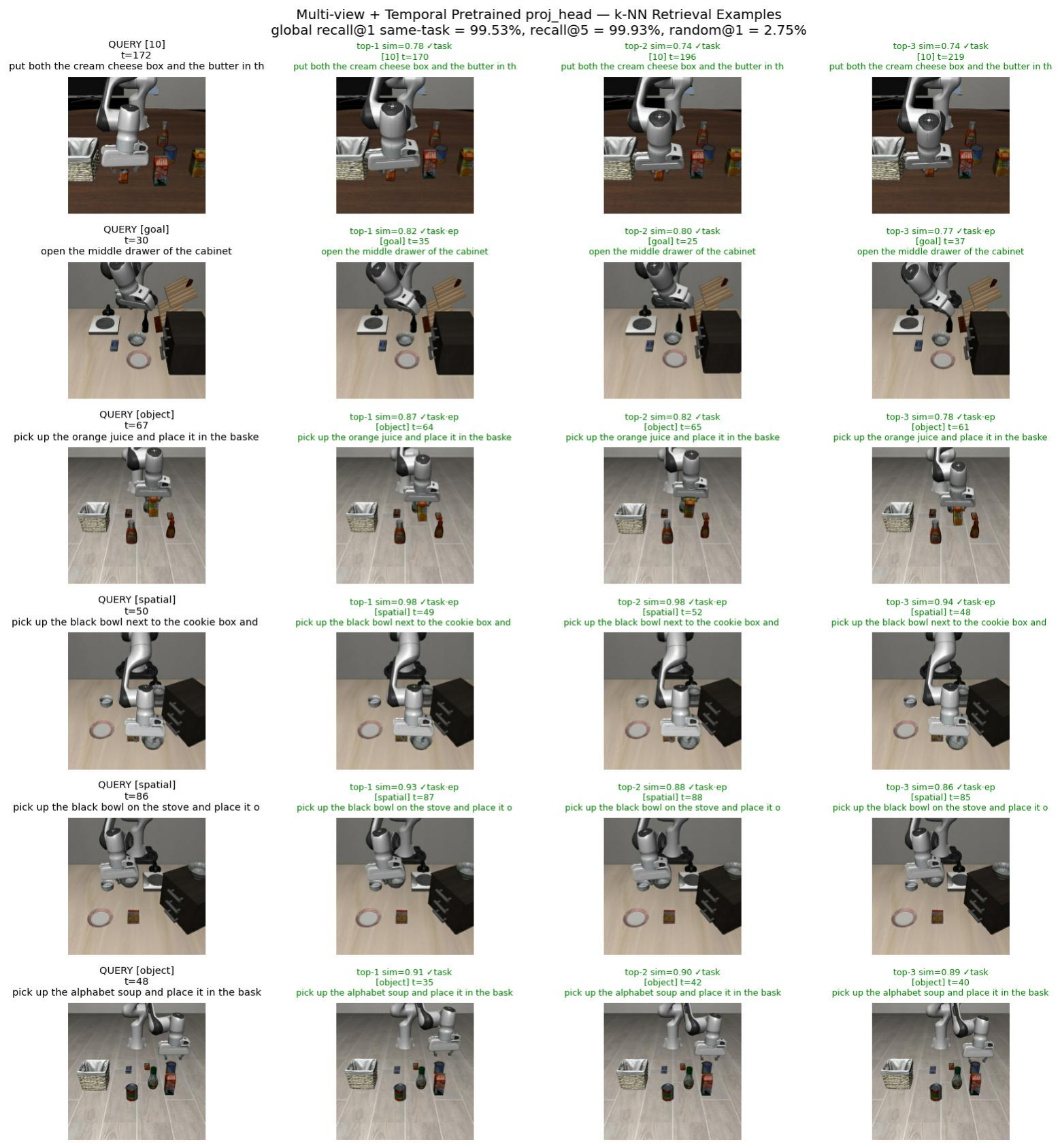}
\caption{Qualitative k-NN retrieval examples: 6 query frames \(\times\) top-3 retrieved neighbours. Ticks indicate same-task hits.}
\end{figure}

The Object suite's perfect 100\% recall@1 aligns with the DoRA Object win in \S\ref{sec:main-result}: Object's tasks have the cleanest task-vs-task visual signal, benefiting both representation learning and parameter-efficient fine-tuning.

\subsubsection*{Downstream Use}

This pretraining produces a \texttt{proj\_head.pt} checkpoint (2.6\,MB, 656K parameters). We have not yet evaluated whether using this as the initial state of OpenVLA's vision encoder during DPO fine-tuning improves downstream success rates---this is a natural next experiment and is noted as future work.

\section{Discussion}

\subsection{What Generalises Across VLA Paradigms}

DPO, with the right log-probability surrogate, \emph{does} transfer from autoregressive to flow-matching VLAs. The training dynamics are qualitatively similar: chosen-vs-rejected reward margin grows monotonically over 500 training steps; the BCE-style loss decreases without instability; the resulting policy preserves SFT-saturated suite performance. \textbf{This is the load-bearing positive result of the paper}: a single post-training pipeline for two architectural lineages of VLA, requiring only a method-level change (closed-form vs surrogate logp) but no protocol-level changes.

\subsection{What Doesn't Generalise: Inference Caching}

KV-cache strategies designed for autoregressive VLAs (VLA-Cache and its near-relatives) do not transfer to flow-matching \(\pi_{0.5}\). The negative result is structural, not implementation-dependent: flow-matching shifts the bulk of inference cost from the prefix forward (where caching helps) into the denoise loop. This produces a clear research direction: \textbf{target the denoise loop}, e.g.\ via consistency-model distillation. Concurrent work SnapFlow~\cite{snapflow2026} validates this prescription empirically.

\subsection{PEFT-Architecture Interaction}

DoRA's gains over SFT are not uniform across LIBERO suites. The pattern correlates with whether the suite demands a genuine direction shift (Spatial: smallest gain, +2.7\,pp) or a magnitude refinement on existing skills (Object +20\,pp, Long-horizon +11\,pp). Combined with DoRA's tighter per-seed spread on Object (zero variance) vs LoRA's 18-pp spread, this suggests \textbf{suite-level diagnosis} as an inexpensive heuristic for adapter choice in production VLA fine-tuning.

\subsection{Implications for Future Work}

\begin{enumerate}
  \item \textbf{Denoise-loop-targeting acceleration} is the productive direction for flow-matching VLA inference. Consistency model distillation~\cite{salimans2022progressive} is a natural fit; SnapFlow~\cite{snapflow2026} provides a validated reference.
  \item \textbf{DoRA on flow-matching backbones} is unstudied. Our work covers OpenVLA only; extending DoRA+DPO to \(\pi_{0.5}\) is straightforward given our pipeline.
  \item \textbf{Multi-view pretraining as VLA encoder init.} Our \texttt{proj\_head.pt} achieves 99.5\% k-NN retrieval but has not been evaluated as a downstream initialisation for VLA fine-tunes.
  \item \textbf{GRPO on flow-matching VLAs.} A full cross-paradigm \(\times\) cross-algorithm grid on richer compute budget (3+ seeds, additional simulators) is a natural full-paper extension.
\end{enumerate}

\section{Limitations}\label{sec:limitations}

\begin{enumerate}
  \item \textbf{Asymmetric multiseed coverage on the LoRA side.} All four DoRA cells are 3-seed pooled (seeds 42, 1337, 2026; 150 trials per suite). On the LoRA side, multiseed (seeds 1337 + 2026) is available for Object/Goal/Long-horizon only. \textbf{LoRA Spatial single-seed only} (s\(=42\) = 78\%), so the DoRA-vs-LoRA Spatial cell in \S\ref{sec:main-result} is reported with the asymmetry footnoted (direct s\(=42\) tie at 78\%; pool comparison shows DoRA 74.7\%). Deferred due to compute budget.
  \item \textbf{No Spirit v1.5 results.} We had originally planned a third backbone (Qwen3-VL flow-matching, distinct from \(\pi_{0.5}\)'s PaliGemma), but its container build path conflicted with our cloudml runtime and we deferred it.
  \item \textbf{GRPO is single-suite.} The reward function design for VLA GRPO was under-explored at our compute budget; our single-suite Spatial GRPO result (App.~C, +2\,pp over SFT) is reported but does not bear the weight of a comparable claim to DPO.
  \item \textbf{\(\pi_{0.5}\) + DPO single-seed only.} The 4-suite \(\pi_{0.5}\) + LoRA + DPO numbers reported in \S\ref{sec:cross-validation} are seed 42 only (50 trials per suite). Multi-seed expansion to match the OpenVLA DoRA 3-seed protocol (\S\ref{sec:main-result}) is left to camera-ready, especially for Long-horizon where the +1.6\,pp gain is the headline cross-paradigm result.
  \item \textbf{Pretrained projection head not evaluated downstream.} \S\ref{sec:pretrain} reports k-NN retrieval as an intrinsic representation quality measure; downstream VLA fine-tune transfer is future work.
  \item \textbf{Sim only.} All evaluations are in LIBERO simulation; real-robot validation is future work.
  \item \textbf{Decoding settings.} OpenVLA SFT baseline is below the OpenVLA paper number on Spatial/Object/Goal, attributable to default temperature 1.0 vs the paper's calibrated decoding. This affects absolute numbers but not the comparative DoRA-vs-SFT finding.
\end{enumerate}

\section{Conclusion}

We presented \textbf{CrossVLA}, an empirical study of cross-paradigm post-training for Vision-Language-Action models. Our contributions: (1)~a surrogate flow-matching log-probability that makes DPO directly applicable to flow-matching VLAs, validated on \(\pi_{0.5}\) across LIBERO 4-suite with mean +1.15\,pp over the openpi paper SFT baseline; (2)~the first DoRA-on-VLA empirical study showing DoRA + DPO improves over OpenVLA SFT by a mean +10.4\,pp across LIBERO 4-suite (600 trials, 3 seeds), with zero seed variance on Object; (3)~a structural negative result on prefix-K/V caching for flow-matching VLAs, with a positive prescription on denoise-loop-targeting acceleration that has been concurrently validated; (4)~a multi-view + temporal contrastive pretraining framework with 99.5\% k-NN recall@1.

The contributions are modest extensions of established techniques to the under-studied flow-matching VLA paradigm. Their value is in the comparative crossing: under one shared protocol, with one open code repository and ckpt set, we map out which post-training techniques transfer cleanly across architectural paradigms and which need paradigm-specific adaptation.

All code, training logs, ckpts, and reproduction scripts are open at \url{https://github.com/lz-googlefycy/vla-lab}.

\bibliographystyle{unsrtnat}
\bibliography{references}

\end{document}